\pdfoutput=1

\documentclass[11pt]{article}

\usepackage{ACL2023}
\usepackage{times}
\usepackage{latexsym}

\usepackage[T1]{fontenc}

\usepackage[utf8]{inputenc}

\usepackage{microtype}

\usepackage{inconsolata}
\usepackage{booktabs}
\usepackage{hyperref}
\usepackage{booktabs} 
\usepackage{CJKutf8}
\usepackage{tabularx}
\usepackage{array}
\usepackage{CJKutf8}
\usepackage{adjustbox}
\usepackage{multirow}
\usepackage{colortbl}
\usepackage{graphicx}
\usepackage[bottom]{footmisc}
\usepackage{lipsum}

\title{MultiPragEval: Multilingual Pragmatic Evaluation \\of Large Language Models}

\author{
    Dojun Park$^{1*}$ \hspace{10pt} 
    Jiwoo Lee$^{1*}$ \hspace{10pt} 
    Hyeyun Jeong$^{1}$ \hspace{10pt} 
    Seohyun Park$^{1}$ \hspace{10pt} \\
    \textbf{Youngeun Koo}$^{2}$ \hspace{10pt} 
    \textbf{Soonha Hwang}$^{3}$ \hspace{10pt} 
    \textbf{Seonwoo Park}$^{1}$ \hspace{10pt} 
    \textbf{Sungeun Lee}$^{1}$ \\
    $^1$Seoul National University, $^2$Sungkyunkwan University, $^3$Yonsei University \\
    \texttt{\{dojun.parkk, seohyun.parkk88\}@gmail.com} \\
    \texttt{\{lee9055, tosirihy, su3503, cristlo5\}@snu.ac.kr} \\
    \texttt{sarah8835@skku.edu}, \texttt{soonha.hwang@yonsei.ac.kr}
}

\begin{document}

\maketitle
\def\thefootnote{*}\footnotetext{These authors contributed equally to this work.}\def\thefootnote{\arabic{footnote}}

\begin{abstract}

As the capabilities of Large Language Models (LLMs) expand, it becomes increasingly important to evaluate them beyond basic knowledge assessment, focusing on higher-level language understanding. This study introduces MultiPragEval, the first multilingual pragmatic evaluation of LLMs, designed for English, German, Korean, and Chinese. Comprising 1200 question units categorized according to Grice's Cooperative Principle and its four conversational maxims, MultiPragEval enables an in-depth assessment of LLMs' contextual awareness and their ability to infer implied meanings. Our findings demonstrate that Claude3-Opus significantly outperforms other models in all tested languages, establishing a state-of-the-art in the field. Among open-source models, Solar-10.7B and Qwen1.5-14B emerge as strong competitors. By analyzing pragmatic inference, we provide valuable insights into the capabilities essential for advanced language comprehension in AI systems. The test suite is publicly available on our GitHub repository at \url{https://github.com/DojunPark/MultiPragEval}.


\end{abstract}

\section{Introduction}

Understanding a language involves not only the ability to process explicit information but also an awareness of the context that influences the meaning of each utterance \citep{sperber1986relevance}. In human communication, context acts as a critical element as it provides a foundation upon which dialogue participants can understand and interact with each other more efficiently. With a shared context, communication becomes more facilitated, allowing subtle nuances to be successfully conveyed, which is essential for engaging in meaningful conversations \citep{krauss1996social}.

With recent advancements in generative AI, current LLMs have demonstrated capabilities that extend far beyond traditional natural language processing (NLP) tasks \citep{brown2020language, achiam2023gpt}. These models are increasingly becoming integral to our daily lives as AI assistants, closely engaging with human users in diverse conversational setups that demand a rapid understanding of the users' needs and intentions, far surpassing mere literal interpretation of text \citep{roller2020recipes}. Given the growing importance of LLMs, accurately evaluating their ability to comprehend context-dependent meanings and demonstrate human-like language comprehension has become crucial \citep{mccoy2019right, xu2020recipes}.

\begin{table}[t]
\begin{tabular}{l|l}
\hline
\textbf{Aspect} & \textbf{Details} \\ \hline
Utterance & \textit{"There's the door."} \\ \hline
\begin{tabular}[c]{@{}l@{}}Literal \\ Meaning\end{tabular} & A door is located over there. \\ \hline
\begin{tabular}[c]{@{}l@{}}Contextual\\ Implication\end{tabular} & \begin{tabular}[c]{@{}l@{}}\textbf{\textit{Context:}} An interviewer says it \\to the interviewee after finishing \\an interview.\\ \textbf{\textit{Implied Meaning:}} The interview \\has concluded and the intervie-\\wee is free to leave the room.\end{tabular} \\ \hline
\end{tabular}
\caption{Literal and contextual implications of the utterance \textit{``There's the door''} in an interview scenario.}
\label{tab:pragmatics}
\end{table}

Pragmatics is a branch of linguistics that studies how language is used to achieve specific goals, where the interpretation of utterances depends not only on their literal meaning but also, crucially, on the surrounding context \citep{grice1975logic}. Consider the example in Table \ref{tab:pragmatics}, which demonstrates both the literal and implied meanings of the utterance, \textit{``There’s the door.''} Literally, this phrase simply indicates the presence of a door in the specified direction. However, from a pragmatic standpoint, it conveys an additional implied meaning in the context of its usage by an interviewer to an interviewee after an interview has concluded. In this scenario, the speaker is subtly suggesting that the interviewee is free to leave the room. This example underscores the critical role that context plays in shaping the interpretation of human language.

Despite the clear need for studies analyzing the pragmatic competence of current LLMs, there is not only a lack of systematic evaluation across various models \citep{chang2024survey} but also a strong bias towards English \citep{guo2023evaluating, bommasani2023holistic}, leaving the pragmatic abilities of LLMs in other languages largely unexplored and difficult to compare. Such oversight demonstrates a significant gap in current evaluation practices, particularly given the multilingual nature of today's state-of-the-art LLMs \citep{kwon2023beyond}.

To address these challenges, our study introduces \textbf{MultiPragEval}, the first multilingual test suite designed for the pragmatic evaluation of LLMs in English, German, Korean, and Chinese. Our suite comprises 300 question units per language, totaling 1200 units. These questions are divided into five categories based on Grice’s Cooperative Principles and the corresponding four conversational maxims: quantity, quality, relation, manner, and an additional category dedicated to assessing mere literal meaning understanding, independent of context.

Our main contributions are as follows:

\begin{itemize}
    \item \textbf{Development of MultiPragEval}: We introduce MultiPragEval, a comprehensive test suite specifically designed to evaluate the pragmatic abilities of LLMs across English, German, Korean, and Chinese. 
    
    \item \textbf{Systematic Evaluation of LLMs}: We conduct a thorough evaluation of 15 state-of-the-art LLMs, including both proprietary and open-source models, assessing their contextual awareness and pragmatic understanding capabilities.
    
    \item \textbf{In-depth Performance Analysis}: We offer a detailed analysis of LLM performance, systematically categorized according to Grice's Cooperative Principle and its maxims, highlighting critical patterns and implications for further enhancements in LLM capabilities.
 
\end{itemize}

\section{Related Work}

\paragraph{Current Practices in LLM Evaluation.}
Benchmarks serve as critical tools for standardized evaluation in the field of LLM studies, enabling fair and systematic comparisons across models trained with diverse architectures and strategies \citep{guo2023evaluating}. These benchmarks span a wide range of domains, from general reasoning \citep{zellers2019hellaswag} to specialized fields such as mathematics \citep{cobbe2021training}, coding \citep{chen2021evaluating}, and biomedical sciences \citep{jin2019pubmedqa}. While comprehensive, they primarily focus on assessing knowledge and logical reasoning, emphasizing explicit semantic meanings over the contextual and implied meanings that can vary in different scenarios \citep{sileo2019pragmatics}.

Leaderboards further enhance the field of LLM evaluation by providing a transparent platform where the performance of various models can directly compete with each other. The Open LLM Leaderboard \citep{open-llm-leaderboard}, featuring a range of rigorous benchmarks, establishes a venue for open-source models to showcase their capabilities, thereby fostering engagement in LLM development among both individual developers and tech companies. Meanwhile, Chatbot Arena \citep{chiang2024chatbot} is gaining recognition as a crowd-sourced evaluation platform. It leverages real-time feedback from users who vote on outputs from two randomly selected models. Models are then ranked on the leaderboard based on their Elo rating \citep{elo1978rating}, thus filling the gaps left by automatic benchmarks.

Recently, efforts have been made to create benchmarks specifically targeted at measuring the capabilities of LLMs in languages such as Chinese \citep{li2023cmmlu} and Korean \citep{son2024kmmlu}. This development contributes to advancing a more inclusive multilingual evaluation landscape.

\begin{table*}[t]
\small
\begin{tabularx}{\textwidth}{@{}cp{0.2\textwidth}p{0.1\textwidth}X@{}}
\toprule
\textbf{Language} & \textbf{Context} & \textbf{Utterance} & \textbf{MCQ} \\
\midrule

English & While visiting Charlie's house, Emily saw a large pile of oranges in the kitchen and asked why there were so many. Charlie responded: & "My uncle lives in Florida." & Choose the most appropriate meaning of the above utterance from the following options. \newline
\textbf{(A)} Charlie's uncle sent the oranges.\newline
(B) Charlie's uncle resides in Florida.\newline
(C) People in Florida do not like oranges.\newline
(D) Charlie's uncle lives in a rural house.\newline
(E) None of the above. \\
\midrule

German & Anna, die Felix besuchte, sah, dass es bei Felix viel Wein gab, und als sie fragte, warum es so viel Wein gab, wie er zu so viel Wein komme, sagte Felix:  & "Mein Onkel betreibt ein Weingut in Freiburg." & Wählen Sie die passendste Bedeutung der obigen Äußerung aus den folgenden Aussagen aus.\newline
\textbf{(A)} Felix hat den Wein von seinem Onkel.\newline
(B) Der Onkel von Felix lebt in Freiburg.\newline
(C) Freiburger lieben keinen Wein.\newline
(D) Der Onkel von Felix wohnt in einem Landhaus.\newline
(E) Keine der obigen Aussagen ist richtig. \\
\midrule

Korean & \begin{CJK}{UTF8}{nanummj}철수 집에 놀러 간 영희는 주방에 많은 귤이 쌓여 있는 것을 보고 귤이 왜 이렇게 많은지 물었고 철수는 다음과 같이 말했다.\end{CJK} & \begin{CJK}{UTF8}{nanummj}"우리 작은 아버지께서 제주도에 사셔."\end{CJK} & \begin{CJK}{UTF8}{nanummj}다음 보기에서 위 발화가 갖는 가장 적절한 의미를 고르세요.\newline
\textbf{(A)} 작은 아버지께서 귤을 보내주었다.\newline
(B) 작은 아버지의 거주지는 제주도이다.\newline
(C) 제주도 사람들은 귤을 좋아하지 않는다.\newline
(D) 작은 아버지께서 전원 주택에 사신다.\newline
(E) 정답 없음.\end{CJK} \\
\midrule

Chinese & \begin{CJK*}{UTF8}{gbsn}王芳去张伟家看到厨房里堆放着几大袋葡萄干，便问为什么有这么多，张伟回答说：\end{CJK*} & \begin{CJK*}{UTF8}{gbsn}"我叔叔住在新疆。"\end{CJK*} & \begin{CJK*}{UTF8}{gbsn}请在以下选项中选择最恰当地表达上述话语含义的选项。\newline
\textbf{(A)} 叔叔给张伟邮了葡萄干。\newline
(B) 张伟的叔叔住在新疆。\newline
(C) 新疆人不喜欢葡萄干。\newline
(D) 张伟的叔叔住在乡间别墅里。\newline
(E) 没有正确答案。\end{CJK*} \\
\bottomrule

\end{tabularx}
\caption{Multilingual test units from the test suite on the maxim of relation, comprising a context, an utterance, and a multiple-choice question (MCQ) to assess the understanding of implied meanings. Charlie’s response indirectly addresses Emily's question, thereby violating the maxim of relation. Assuming adherence to the cooperative principle, the most appropriate interpretation is option (A), indicating that Charlie's uncle sent the oranges.}
\label{tab:multilingual}
\end{table*}

\paragraph{Pragmatic Evaluation of LLMs.}
As LLMs continue to evolve, it has become crucial to evaluate how effectively they consider context, which crucially shapes meanings beyond their literal interpretations. \citet{bojic2023gpt} examined multiple LLMs under the framework of Grice’s Cooperative Principle and its conversational maxims to assess their capabilities in understanding implicature. The results demonstrated that GPT-4 \citep{achiam2023gpt} outperformed other models, including human performance. However, the human participants were not native English speakers but educated individuals from Serbia, which potentially limits the impact of the findings.

\citet{di2023pragmatic} conducted a comparable study focusing on GPT-3.5, leveraging the APACS test set \citep{arcara2016test}, which consists of various subtasks such as interviews, descriptions, and narratives. The tests were conducted in both English and Italian, with results reported for Italian due to no notable differences between the two. The findings indicate that GPT-3.5 comes close to human ability but reveals weaknesses in understanding physical metaphors and jokes.

Focusing on Korean, \citet{park2024pragmatic} employed 120 test questions aligned with the four Gricean maxims to further probe the capabilities of various LLMs. The findings demonstrate that GPT-4 excelled in both multiple-choice and open-ended question setups, with HyperCLOVA X \citep{yoo2024hyperclova}, a Korean-specific LLM, closely following. The study also explored in-context learning, demonstrating that the few-shot learning technique consistently leads to positive outcomes across all tested models.

\citet{sravanthi2024pub} introduce a comprehensive pragmatic benchmark that evaluates LLMs across 14 distinct tasks, including implicature, presupposition and deictic detection. Comprising 28k data points, this benchmark aims to provide a nuanced assessment of LLMs' pragmatic abilities, marking a substantial contribution to the field. Yet, there remains a significant need to extend these evaluations to multiple languages to thoroughly assess the multilingual capabilities of LLMs.

\begin{table*}[t]
\small
\centering
\begin{tabular}{@{}lp{6cm}p{7cm}@{}}
\toprule
\textbf{Maxim} & \textbf{Description} & \textbf{Specific Cases Covered} \\ \midrule
Quantity & Make your contribution as informative as is required. & Tautology, insufficient information, excessive information, and cases where the maxim is abided by. \\
Quality & Try to make your contribution one that is true. & Irony, hyperbole, and misinformation. \\
Relation & Ensure that all the information you provide is relevant to the current conversation. & Unrelated information and cases where the maxim is abided by.  \\
Manner & Be perspicuous; Be brief and orderly, and avoid obscurity and ambiguity. & Ambiguity, vagueness, double negation, verbosity, improper order, complicated expressions, and cases where the maxim is abided by. \\ \bottomrule
\end{tabular}
\caption{Grice's maxims and their principles with related linguistic phenomena}
\label{tab:maxims}
\end{table*}

\section{Methodology}

\subsection{Theoretical Foundations of Pragmatics}

To accurately assess the contextual awareness of LLMs, we primarily focus on implicature, based on Grice’s theory \citep{grice1975logic}. Implicature refers to a specific way language is used, in which the literal meaning of an utterance differs from the intended meaning of the speaker, requiring the listener to infer the intended meaning from the surrounding context. This concept is critical for evaluating how well LLMs understand human language, particularly in their ability to capture nuanced meanings beyond the explicit words.

Grice introduced the Cooperative Principle that explains how speakers and listeners cooperate to achieve mutual understanding, and its four conversational maxims, which suggest how an utterance should desirably be conducted. Detailed in Table \ref{tab:maxims}, the maxim of quantity requires information to be as informative as necessary--neither more nor less. The maxim of quality emphasizes the importance of offering truthful contributions. The maxim of relation ensures all information is pertinent to the current conversation. The maxim of manner demands clarity and brevity, avoiding obscurity and ambiguity.

Considering the critical role of understanding implicated meanings in communication, this study investigates LLMs' comprehension of conversational implicatures. Specifically, we evaluate LLMs' capabilities in inferring implied meanings that arise from either abiding by or violating these maxims.

\subsection{Development of the Test Suite}

To develop our test suite, we followed a structured process divided into three key phases: describing the initial dataset, expanding its scope, and translating it into the target languages and verifying the translations. Table \ref{tab:multilingual} showcases an example of a test unit focused on the maxim of relation from our complete test suite, presented in English, German, Korean, and Chinese.

\paragraph{Initial Dataset.} The development of the MultiPragEval test suite began with the foundational work by \citep{park2024pragmatic}, who crafted a set of 120 question units designed to assess LLMs in terms of four conversational maxims. Each maxim was represented by 30 units, which included a structured scenario setting the conversational context, an utterance by a participant, and a set of questions comprising both a multiple-choice question and an open-ended question. We adopted the context, utterance, and multiple-choice question components from this test set as our starting point.

\paragraph{Expansion.} Next, we expanded the number of question units from 120 to 300 to encompass a broader range of pragmatic contexts. Each conversational maxim, originally represented by 30 units, was doubled to 60 to deepen the evaluative scope, including more diverse linguistic phenomena as shown in Table \ref{tab:maxims}. Additionally, we introduced a new category specifically designed to assess the understanding of literal meanings, which allows us to explore potential trade-offs between performances in understanding literal versus implied meanings. To further enhance the complexity of our test suite, we included units that do not have a correct answer by adding a `None of the above' option to the multiple-choice setups.

\paragraph{Translation and Verification.} In the subsequent phase, we translated the Korean test set into English, German, and Chinese using DeepL \footnote{\url{https://www.deepl.com}} for the initial conversion. Then, Korean-native linguistic experts with CEFR C1 \footnote{\url{https://www.coe.int/en/web/common-european-framework-reference-languages/level-descriptions}} level proficiency in the target languages refined the translations to ensure that these translations preserved the intended meanings and nuances. They also adapted cultural elements by substituting the names of characters and setting details to reflect the local context of each language. Finally, native speakers of each target language, who hold degrees in linguistics and related fields, conducted a thorough verification of the translations. This process confirmed that the quality and accuracy of the translations were on par with the original Korean versions.

\subsection{Experimental Setup}

\begin{table}[t]
\centering
\begin{tabular}{@{}lll@{}}
\toprule
\textbf{Type} & \textbf{Model} & \textbf{Version} \\ \midrule
\multirow{8}{*}{Proprietary} & GPT-3.5 & turbo-0125 \\
 & GPT-4 & turbo-2024-04-09 \\
 & Claude3-Haiku & haiku-20240307 \\
 & Claude3-Sonnet & sonnet-20240229 \\
 & Claude3-Opus & opus-20240229 \\
 & Mistral-small & small-2402 \\
 & Mistral-medium & medium-2312 \\
 & Mistral-large & large-2402 \\ \midrule
\multirow{7}{*}{Open-Src.} 
 & Llama-2-13B & chat-hf \\
 & Llama-2-7B & chat-hf \\
 & Llama-3-8B & Instruct \\
 & Gemma-7B & 1.1-7b-it \\
 & Solar-10.7B & Instruct-v1.0 \\
 & Qwen-14B & 1.5-14B-Chat \\
 & Qwen-7B & 1.5-7B-Chat \\ \bottomrule
\end{tabular}
\caption{Overview of proprietary and open-source LLMs evaluated in the study}
\label{tab:models}
\end{table}

\begin{table*}[t]
\centering
\begin{tabular}{@{}lcccccccc@{}}
\toprule
 & \multicolumn{5}{c}{\textbf{English}} & \multicolumn{1}{c}{} & \multicolumn{1}{c}{} & \multicolumn{1}{c}{} \\ \cmidrule(r){1-6}
 & \multicolumn{1}{l}{Quan.} & \multicolumn{1}{l}{Qual.} & \multicolumn{1}{l}{Rel.} & \multicolumn{1}{l}{Man.} & \multicolumn{1}{l}{\textbf{Avg.}} & \multicolumn{1}{c}{\multirow{-2}{*}{\textbf{\begin{tabular}[c]{@{}c@{}}German\\ Avg.\end{tabular}}}} & \multicolumn{1}{c}{\multirow{-2}{*}{\textbf{\begin{tabular}[c]{@{}c@{}}Korean\\ Avg.\end{tabular}}}} & \multicolumn{1}{c}{\multirow{-2}{*}{\textbf{\begin{tabular}[c]{@{}c@{}}Chinese\\ Avg.\end{tabular}}}} \\ \midrule
GPT-4 & \cellcolor[HTML]{ECF4FF}65.00 & \cellcolor[HTML]{3399FF}83.89 & \cellcolor[HTML]{34CDF9}82.22 & \cellcolor[HTML]{DAE8FC}70.00 & \textbf{75.28} & \textbf{72.50} & \textbf{81.25} & \textbf{68.75} \\
GPT-3.5 & \cellcolor[HTML]{DAE8FC}51.11 & \cellcolor[HTML]{3399FF}66.67 & \cellcolor[HTML]{34CDF9}52.78 & \cellcolor[HTML]{ECF4FF}42.89 & 53.61 & 52.92 & 38.89 & 43.61 \\
Claude3-Opus & \cellcolor[HTML]{DAE8FC}81.11 & \cellcolor[HTML]{3399FF}88.89 & \cellcolor[HTML]{3399FF}88.89 & \cellcolor[HTML]{DAE8FC}81.11 & \textbf{85.00} & \textbf{82.78} & \textbf{87.08} & \textbf{76.67} \\
Claude3-Sonnet & \cellcolor[HTML]{DAE8FC}62.22 & \cellcolor[HTML]{3399FF}81.67 & \cellcolor[HTML]{34CDF9}67.22 & \cellcolor[HTML]{ECF4FF}54.44 & \underline{66.39} & \underline{60.14} & \underline{63.33} & \underline{48.61} \\
Claude3-Haiku & \cellcolor[HTML]{DAE8FC}56.67 & \cellcolor[HTML]{3399FF}67.78 & \cellcolor[HTML]{34CDF9}58.89 & \cellcolor[HTML]{ECF4FF}43.33 & 56.67 & 45.14 & 38.47 & 40.83 \\
Mistral-Large & \cellcolor[HTML]{34CDF9}61.11 & \cellcolor[HTML]{3399FF}71.11 & \cellcolor[HTML]{34CDF9}61.11 & \cellcolor[HTML]{ECF4FF}52.22 & \underline{61.39} & \underline{63.75} & \underline{65.56} & \underline{54.72} \\
Mistral-Medium & \cellcolor[HTML]{ECF4FF}61.11 & \cellcolor[HTML]{34CDF9}69.44 & \cellcolor[HTML]{3399FF}72.22 & \cellcolor[HTML]{DAE8FC}62.22 & \underline{66.25} & 53.61 & 52.92 & 38.89 \\
Mistral-Small & \cellcolor[HTML]{34CDF9}57.22 & \cellcolor[HTML]{3399FF}57.78 & \cellcolor[HTML]{DAE8FC}54.44 & \cellcolor[HTML]{ECF4FF}35.00 & 51.11 & 51.11 & 40.42 & 33.61 \\ \midrule
Llama3-8B & \cellcolor[HTML]{34CDF9}54.44 & \cellcolor[HTML]{3399FF}68.89 & \cellcolor[HTML]{ECF4FF}44.44 & \cellcolor[HTML]{DAE8FC}45.56 & 53.33 & 40.00 & 32.50 & 46.81 \\
Llama2-13B & \cellcolor[HTML]{DAE8FC}26.67 & \cellcolor[HTML]{3399FF}32.22 & \cellcolor[HTML]{ECF4FF}16.67 & \cellcolor[HTML]{3399FF}32.22 & 26.94 & 16.39 & \textbf{47.50} & \underline{8.75} \\
Llama2-7B & \cellcolor[HTML]{3399FF}31.11 & \cellcolor[HTML]{34CDF9}26.67 & \cellcolor[HTML]{ECF4FF}11.11 & \cellcolor[HTML]{DAE8FC}18.33 & 21.81 & \underline{4.44} & \underline{3.06} & \underline{4.17} \\
Gemma-7B & \cellcolor[HTML]{3399FF}37.78 & \cellcolor[HTML]{34CDF9}36.67 & \cellcolor[HTML]{DAE8FC}35.00 & \cellcolor[HTML]{ECF4FF}30.56 & 35.00 & 27.22 & 20.83 & 25.28 \\
Solar-10.7B & \cellcolor[HTML]{DAE8FC}58.33 & \cellcolor[HTML]{3399FF}65.56 & \cellcolor[HTML]{34CDF9}62.22 & \cellcolor[HTML]{ECF4FF}51.11 & \textbf{59.31} & \textbf{55.69} & \textbf{49.03} & 46.39 \\
Qwen-14B & \cellcolor[HTML]{DAE8FC}52.22 & \cellcolor[HTML]{3399FF}61.67 & \cellcolor[HTML]{34CDF9}56.11 & \cellcolor[HTML]{ECF4FF}43.33 & 53.33 & 43.06 & \textbf{49.72} & \textbf{50.00} \\
Qwen-7B & \cellcolor[HTML]{34CDF9}53.89 & \cellcolor[HTML]{3399FF}62.22 & \cellcolor[HTML]{DAE8FC}47.22 & \cellcolor[HTML]{ECF4FF}37.78 & 50.28 & 39.44 & 35.14 & 41.11 \\ \bottomrule
\end{tabular}
\caption{Performance of LLMs on the MultiPragEval test suite: scores across four languages and by maxims with overall averages; Leading scores among proprietary and open-source models are highlighted in bold. The scores for each maxim are color-coded in shades of blue to represent the relative ranking within each model.}
\label{tab:score}
\end{table*}

\paragraph{Models.} Our study includes 15 LLMs, categorized into two types: proprietary LLMs accessed via API, and open-source LLMs where we have direct access to the model weights. As detailed in Table \ref{tab:models}, the proprietary models comprise two GPT models \citep{achiam2023gpt} by OpenAI, along with three different sizes of both Claude3 \citep{anthropic2024claude} by Anthropic and Mistral by Mistral AI \footnote{\url{https://mistral.ai/}}. We exclude Gemini by Google from our analysis due to its limited accessibility via API.

Additionally, we evaluate publicly available open-source models, each with approximately 10 billion parameters. These models were selected based on two criteria: their architecture (Transformer decoder-based models) and their performance on publicly accessible benchmarks. The selected models include three Llama models \citep{touvron2023llama} by Meta, Gemma \citep{team2024gemma} by Google, Solar \citep{kim2023solar} by Korean company Upstage, and two Qwen models \citep{bai2023qwen} by Chinese firm Alibaba, with consideration also given to the diversity of languages represented in our study.

\paragraph{LLM Response Generation.} To generate answers from each LLM, we set the temperature hyperparameter at 0.5 across models to balance coherence and creativity in their responses. For inference on the open-source LLMs, we utilized a single H100-80GB unit. Each model was queried three times to account for the inherent randomness in responses. We then computed the average score for each model across these trials to ensure a robust assessment of performance for each LLM iteration. Scores were calculated based on the ratio of correct answers to the total number of test units across all three trials. The actual prompt for the experiment and inter-rater agreement across three trials are detailed in the Appendix \ref{sec:append_b}.

\section{Result}

\subsection{Analysis of LLM Performance}

\paragraph{Overall Performance.}
Table \ref{tab:score} presents the results from the evaluation of the selected LLMs on the MultiPragEval test suite. It demonstrates that Claude3-Opus significantly outperforms all other models across four languages, with GPT-4 trailing by approximately 6-10 points. This performance gap underscores Claude3-Opus's exceptional ability to capture the subtle nuances of language that are highly context-dependent. These findings highlight its position as the most proficient among the current state-of-the-art LLMs across English, German, Korean, and Chinese.

Mistral-Large and Claude3-Sonnet are closely matched for the next tier of performance; Mistral-Large outperforms Claude3-Sonnet in German, Korean, and Chinese. However, Claude3-Sonnet achieves a higher score in English, registering 66.39 compared to Mistral-Large's 61.39. Interestingly, while Mistral-Large generally shows improved scores across languages compared to Mistral-Medium, it scores lower in English, dropping to 61.39 from the medium-sized model's 66.25.

Solar-10.7B demonstrates stable performance, consistently outperforming GPT-3.5 across all four languages. It is the only open-source model that surpasses GPT-3.5 in both English and German. In English, it closely follows Mistral-Large with a score of 59.31 and is just behind Claude3-Sonnet in German, with a score of 55.69.

Qwen-14B also stands out among other open-source LLMs, outperforming its counterparts with scores of 50.00 in Chinese and 49.72 in Korean. In contrast, both Llama2-13B and Llama2-7B demonstrate a strong bias towards literal interpretations yielding poor scores, while Llama3-8B shows enhanced performance compared to its earlier versions. Notably, Llama2-13B achieves a significant leap in Korean, scoring 47.50 compared to Llama2-7B's 3.06, while exhibiting a more gradual increase in other languages.

\paragraph{Performance Gap Across Languages. } 
We observed that the models generally achieve higher performance scores in English than in other languages, likely due to larger English training datasets enhancing reasoning capabilities. Interestingly, flagship proprietary models like GPT-4, Claude-Opus, and Mistral-large show slightly better performance in Korean. We believe there could be two possible reasons for this performance gap. First, it is possible that the initial Korean dataset, from which we extended our test suite \citep{park2024pragmatic}, was used in model training, allowing the models to better understand newly created Korean questions that follow the same template. Secondly, the gap could stem from the test suite being initially developed in Korean and then translated into other languages. Cultural nuances and conventions embedded in each language may lead to subtle differences in how the same expressions are interpreted, with the implications being understood differently depending on the language region.

Significant performance discrepancies were also observed across models. Claude-Haiku scored 56.7 in English but only 38.4 in Korean, while Mistral-small dropped from 51.1 in English to 33.6 in Chinese. Llama2-13B showed the largest gap, with scores of 47.5 in Korean versus 8.7 in Chinese. These differences highlight language-specific biases in the models, indicating a need for improvements to boost multilingual capabilities.

\begin{figure*}[t!]
    \centering
    \includegraphics[width=\textwidth, page=1]{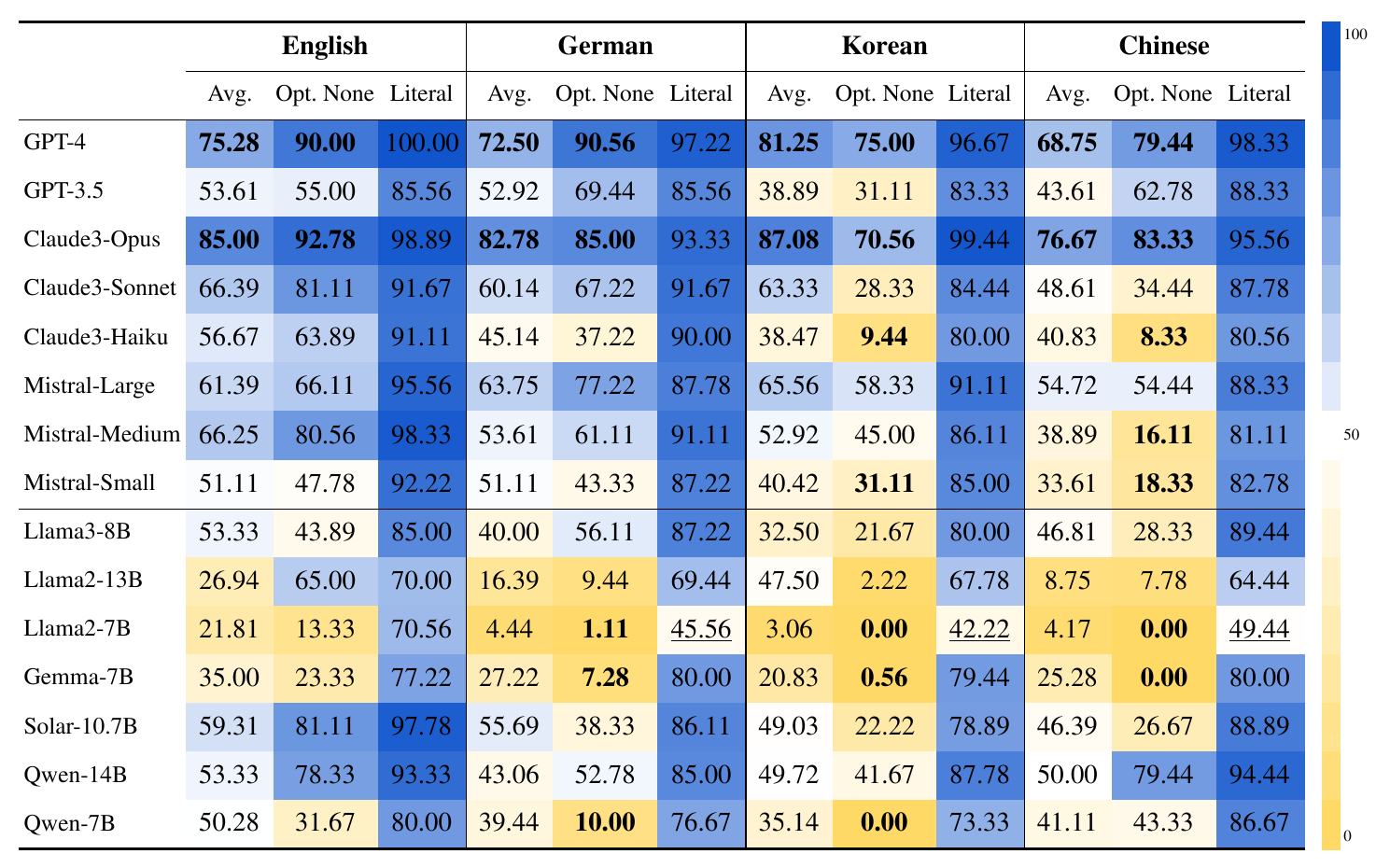}
    \caption{Breakdown of LLM scores for `No Correct Answers' and literal meaning tests across four languages; the heatmap uses two colors--blue indicating higher scores and yellow indicating lower scores.}
    \label{fig:score}
\end{figure*}

\paragraph{Closer Look at Individual Maxims.}
Table \ref{tab:score} also shows the performance scores of LLMs on individual maxims in the English test suite. We observe a consistent pattern across LLMs where scores for the maxim of quality generally rank highest, while scores for the maxim of manner rank lowest. This pattern is not unique to English but is also observable in other languages, suggesting a universal trend (see Appendix \ref{sec:appendix}). This outcome is expected because expressions governed by the maxim of quality, which become untrue statements when interpreted literally, make it easier for LLMs to infer the appropriate implied meanings. Conversely, the maxim of manner, involving verbose or ambiguous expressions, poses more subtle challenges that likewise pose difficulties for humans \citep{hoffmann2010sprachwissenschaft}.

Another noteworthy observation is that as the overall performance increases, the scores for the maxim of relation also generally improve. This pattern is more evident among proprietary models, where the maxim of relation mostly ranks second. Similarly, Solar-10.7B and Qwen-14B, which perform comparably to GPT-3.5, achieve higher scores in the maxim of relation compared to those of quantity and manner. Conversely, other open-source models with lower average scores tend to have lower rankings in the maxim of relation, falling below the maxim of quantity. This suggests that capturing relevancy within the given context plays a significant role in a more precise interpretation of implied information, contributing to better overall performance.

\begin{table*}[t]
\small
\centering
\begin{tabular}{@{}lccccccc@{}}
\toprule
\textbf{Model} & \multicolumn{1}{l}{\begin{tabular}[c]{@{}c@{}}\textbf{MultiPragEval}\\ (Eng.)\end{tabular}} & \multicolumn{1}{l}{\begin{tabular}[c]{@{}c@{}}\textbf{MMLU}\\ 5-shot\end{tabular}} & \multicolumn{1}{l}{\begin{tabular}[c]{@{}c@{}}\textbf{MATH}\\ 4-shot\end{tabular}} & \textbf{Arena Elo}\textsuperscript{*} & \multicolumn{1}{l}{\begin{tabular}[c]{@{}c@{}}\textbf{ARC}\\ 25-shot\end{tabular}} & \multicolumn{1}{l}{\begin{tabular}[c]{@{}c@{}}\textbf{HumanEval}\\ 0-shot\end{tabular}} & \multicolumn{1}{l}{\begin{tabular}[c]{@{}c@{}}\textbf{GSM-8K}\\ 8-shot\end{tabular}} \\ \midrule
GPT-4 & \textbf{75.28} & \underline{86.4} & \textbf{52.9} & 1252 & 96.3 & 67.0 & 92.0 \\
GPT-3.5 & 53.6 & 70.0 & 34.1 & 1110 & 85.2 & 48.1 & 57.1 \\
Claude3-Opus & \textbf{85.0} & \underline{86.8} & \textbf{61.0} & 1246 & 96.4 & 84.9 & 95.0 \\
Claude3-Sonnet & 66.4 & 79.0 & 40.5 & 1199 & 93.2 & 73.0 & 92.3 \\
Claude3-Haiku & 56.7 & 75.2 & 40.9 & 1181 & 89.2 & 75.9 & 88.9 \\
Llama3-8B & 53.3 & 68.4 & 30.0 & 1154 & 60.7 & 62.2 & 79.6 \\
Llama2-13B & 26.9 & 47.8 & 6.7 & 1065 & 59.4 & 14.0 & 77.4 \\
Llama2-7B & 21.8 & 34.1 & 3.8 & 1042 & 53.1 & 7.9 & 25.7 \\
Gemma-7B & 35.0 & 66.0 & 24.3 & 1091 & 61.1 & 32.3 & 46.4 \\
Qwen-14B & 53.3 & 69.4 & 24.8 & 1119 & 56.6 & 32.3 & 61.3 \\
Qwen-7B & 50.3 & 61.7 & 11.6 & 1079 & 54.2 & 29.9 & 51.7 \\ \midrule
\textbf{Kendall $\tau$} & \textbf{1.00} & \textbf{0.95} & \textbf{0.92} & \textbf{0.84} & \textbf{0.81} & \textbf{0.80} & \textbf{0.73} \\ \bottomrule
\end{tabular}
\caption{Performance scores of LLMs across multiple benchmarks and Kendall’s Tau correlation Coefficients Relative to MultiPragEval.\\\textsuperscript{*} The Arena Elo scores are as of May 17, 2024.}
\label{tab:correlation}
\end{table*}

\subsection{Assessing the Stability of Pragmatic Inference}

We further explore the stability of LLMs in pragmatic inference under two specific setups. First, we evaluate the models on a subset of each category of maxims, specifically designed where the test questions lack an appropriate answer. This subset is intended to be more challenging as it requires the models to identify incorrect interpretations and select the option `(E) None of the above' without reference to a correct meaning. Secondly, we test the models on additional test units consisting of context, utterance, and question, structured similarly, but where the context is irrelevant to the utterance. This setup is designed to assess whether LLMs can accurately distinguish purely literal meanings from inappropriate interpretations.

\paragraph{Subset of No Correct Answer.} Figure \ref{fig:score} illustrates that the scores on the subset without correct answers (Opt. None) generally align with the overall scores, yet they reveal subtle differences in performance details. While Claude3-Opus consistently outperforms GPT-4 by a certain margin in overall scores across all languages, GPT-4 surpasses Claude3-Opus by approximately 5 points in both German and Korean. This result indicates that both models are comparably robust in the challenging setup of pragmatic consideration.

It is evident that models with lower overall scores exhibit significant declines when tested in the setup without a correct answer. Among proprietary LLMs, Claude3-Haiku, along with medium and small-sized models by Mistral, notably drop in scores, indicating their struggles with the task. Similarly, 7-billion parameter models such as Llama2, Gemma, and Qwen also show poor performance, underscoring the complexity of the task for models of this size.

\paragraph{Additional Set of Literal Meaning.} The scores on the set asking literal meanings also demonstrate a general increase along with the overall scores. While the flagship models of GPT and Claude show performance close to perfect, GPT-4 demonstrates a slight edge over Claude-3-Opus for English, German, and Chinese. This may suggest a trade-off between pragmatic and literal focus in their inferences.

The Llama2 models, particularly Llama2-7B, show the lowest scores among the others, with 42.22, 45.56, and 49.44 for Korean, German, and English, respectively. These results generally correlate with lower overall scores in both the pragmatic and no-correct-answer subset questions. We interpret this to mean that these tasks are not independent of each other, but instead mutually influence one another, highlighting the importance of maintaining a good balance between the sub-tasks.

\subsection{Comparison with Existing Benchmarks}

To further delve into the implications of our findings, we compare the results from our English test suite with existing English-based benchmarks. This analysis encompasses scores from 11 models, for which other benchmark scores were publicly available. We consider seven popular benchmarks: MMLU \citep{hendrycks2020measuring} and ARC \citep{clark2018think} for general reasoning, HumanEval \citep{chen2021evaluating} for coding, GSM-8K \citep{cobbe2021training} and MATH \citep{hendrycks2021measuring} for mathematics, and Chatbot Arena \citep{chiang2024chatbot}, a crowd-sourced evaluation. We opted to calculate the correlation coefficients using Kendall’s Tau \citep{kendall1938new} due to its better handling of varying ranges and subtle differences between benchmarks.

The correlations of MultiPragEval with other benchmarks consistently show high values, indicating a general trend toward `good' performance across different benchmarks. This suggests that improvements in a model's performance on one task generally enhance its performance on other tasks \citep{raffel2020exploring}.

MMLU and MATH exhibit the highest correlations among other benchmarks, suggesting that the abilities assessed by these benchmarks align closely with those required for pragmatic inference. It is anticipated that MMLU, which evaluates the general language understanding capabilities of LLMs across a broad spectrum of disciplines, reflects the ability to consider contextual information in language, which is a key requirement of MultiPragEval.

However, the high correlation observed with the MATH benchmark is surprising, given its primary focus on mathematical reasoning. Notably, the score gap between Claude3-Opus and GPT-4, which is around 10 points on MultiPragEval, is similarly reflected on MATH but not distinctively on MMLU. This pattern suggests that the sophisticated mathematical problem-solving required by MATH--which demands a higher level of logical reasoning compared to the basic mathematical problems in GSM-8K--may also tap into core capabilities essential for pragmatic inference. This connection between mathematical reasoning and high-level linguistic comprehension indicates an intricate relationship that requires deeper investigation.

\section{Conclusion}

In this work, we present the first multilingual study of LLMs’ capabilities of their pragmatic inference, particularly in the context of Grice’s theory of conversational implicature. Our findings demonstrate the usefulness of MultiPragEval test suite in distinguishing the levels of comprehension among various proprietary and open-source models.

The results reveal that among the models evaluated, Claude3-Opus and GPT-4 particularly stand out, with Claude3-Opus consistently outperforming GPT-4 by 6 to 10 points across all languages, affirming its state-of-the-art capability in pragmatic understanding. Top-performing open-source models like Solar-10.7B and Qwen-14B demonstrate superior or comparable performance to lite-size proprietary models such as GPT-3.5, Claude3-Haiku, and Mistral-Small. The performance gaps across languages within models and individual Grice’s maxims further highlight language biases and areas for improvement. 

Our findings, with the highest correlations with MMLU and MATH, suggest that general language understanding and complex logical reasoning are intricately linked to pragmatic inference abilities. This insight guides us towards further research to empirically demonstrate how these abilities relate to pragmatic reasoning.

\section*{Limitations}

While our study provides a comprehensive comparison of 15 proprietary and open-source models, it does not include a comparison with human performance. Including human performance would offer deeper insights into how closely LLMs approximate human abilities. Moreover, human performance can vary across languages, which would enrich our understanding of the LLMs’ multilingual pragmatic abilities. Recognizing this gap, we aim to incorporate human performance comparisons in our future research.

Another limitation of our study is its exclusive focus on implicature, despite pragmatics encompassing a broader range of phenomena such as speech acts, presupposition, and politeness. This focus was chosen due to the increasing role of LLMs as AI assistants, which often need to interpret human expressions that are frequently conveyed implicitly. The ability of LLMs to capture these subtle nuances directly influences human judgments about the quality of these systems. Furthermore, contextual awareness is critical not only for linguists but also for NLP engineers who aim to provide reliable services to users. We believe that our specific focus on implicature provides valuable insights into how effectively current LLMs manage the complexities inherent in interpreting implied meanings, a crucial aspect of human communication.

Our study set the temperature value to 0.5 to achieve a moderate balance between consistency and creativity in responses. However, it is important to note that the optimal temperature may vary for each LLM, and the effect of temperature settings on pragmatic inference remains unclear. Recognizing the potential influence of temperature on LLMs' pragmatic abilities, we suggest that future studies investigate the relationship between temperature and pragmatic reasoning to gain deeper insights into how LLMs handle nuanced language tasks.

\section*{Ethics Statement}

In this work, we introduce a test suite designed to evaluate the pragmatic abilities of LLMs. We have ensured that all data created for this study does not infringe on any existing intellectual property rights, while also ensuring it contains no personally identifiable information. Linguistic experts were involved in the creation and translation of the test suite; all contributors were fully informed about the research's purpose and the methods employed. We commit to making the dataset publicly available to foster transparency and further research in the field.

\section*{Acknowledgements}
This research was supported by Basic Science Research Program through the National Research Foundation of Korea (NRF) funded by the Ministry of Education (RS-2023-00274280). Additionally, this work was supported by Artificial intelligence industrial convergence cluster development project funded by the Ministry of Science and ICT(MSIT, Korea)\&Gwangju Metropolitan City.

\bibliography{anthology,custom}
\bibliographystyle{acl_natbib}

\clearpage
\appendix
\onecolumn

\section{Demonstration of Test Unit Example}
\label{sec:appendix}


\vspace*{\fill}

\begin{table}[h]
\small
\begin{tabularx}{\textwidth}{@{}cp{0.2\textwidth}p{0.1\textwidth}X@{}}
\toprule
\textbf{Language} & \textbf{Context} & \textbf{Utterance} & \textbf{MCQ} \\
\midrule

English & A student asks their professor if they can extend the due date of an assignment just a little longer. The professor relies: & "Rules are rules." & Choose the most appropriate meaning of the above utterance from the following options.\newline
\textbf{(A)} The deadline can't be extended because rules must be followed.\newline
(B) Rules are rules.\newline
(C) Breaking the rules isn't a big deal, so I'll give the student a chance.\newline
(D) The professor discovered a new theory after doing research.\newline
(E) None of the above. \\

\midrule

German & Ein Student fragt seinen Professor, ob er den Abgabetermin für eine Aufgabe noch ein wenig hinauszögern kann.  & "Regeln sind Regeln." & Wählen Sie die passendste Bedeutung der obigen Äußerung aus den folgenden Aussagen aus.\newline
\textbf{(A)} Die Frist kann nicht verlängert werden, weil die Regeln eingehalten werden müssen.\newline
(B) Regeln sind Regeln.\newline
(C) Ein Verstoß gegen die Regeln ist keine große Sache, also gebe ich dem Studenten eine Chance.\newline
(D) Der Professor hat durch Nachforschungen eine neue Theorie entdeckt.\newline
(E) Keine der obigen Aussagen ist richtig.\\
\midrule

Korean & \begin{CJK}{UTF8}{nanummj}학생이 교수에게 과제의 마감 기한을 조금만 더 늘려 주실 수 없냐고 부탁하자 교수가 말한다.\end{CJK} & \begin{CJK}{UTF8}{nanummj}"규칙은 규칙일세."\end{CJK} & \begin{CJK}{UTF8}{nanummj}다음 보기에서 위 발화가 갖는 가장 적절한 의미를 고르세요.\newline
\textbf{(A)} 규칙은 지켜져야만 하므로 마감 기한을 늘릴 수 없다.\newline
(B) 규칙은 규칙이다. \newline
(C) 규칙을 깨는 것은 큰 문제가 되지 않으므로 학생에게 기회를 주겠다. \newline
(D) 교수는 연구 끝에 새로운 이론을 발견했다.\newline
(E) 정답 없음.\end{CJK} \\
\midrule

Chinese & \begin{CJK*}{UTF8}{gbsn}一名学生问教授可不可以将作业的截止日期再延长一点，教授说:\end{CJK*} & \begin{CJK*}{UTF8}{gbsn}"规则就是规则。"\end{CJK*} & \begin{CJK*}{UTF8}{gbsn}请在以下选项中选择最恰当地表达上述话语含义的选项。\newline
\textbf{(A)} 规则必须遵守，因此不能延长截止期限。\newline
(B) 规矩就是规矩。\newline
(C) 违反规则没什么大不了的，所以教授会给学生一个机会。\newline
(D) 教授经过研究发现了一个新理论。\newline
(E) 没有正确答案。\end{CJK*} \\
\bottomrule

\end{tabularx}
\caption{Multilingual test unit example on the maxim of quantity. The utterance "Rules are rules" is not sufficiently informative because it provides less information than necessary. This under-informativeness constitutes a violation of Grice's maxim of quantity, which demands that enough information be given to be fully informative. In this context, "the rules" implicitly refer to the adherence to established guidelines, such as the due date for assignments. Therefore, the most appropriate interpretation of the professor's statement is option (A) "The deadline can't be extended because rules must be followed," which accurately captures the implied meaning behind the response.}
\label{tab:quan}
\end{table}


\vspace*{\fill}

\begin{table}[h]
\small
\begin{tabularx}{\textwidth}{@{}cp{0.2\textwidth}p{0.1\textwidth}X@{}}
\toprule
\textbf{Language} & \textbf{Context} & \textbf{Utterance} & \textbf{MCQ} \\
\midrule

English & When Emily, a PhD student, spoke at length about the theory she had studied yesterday, Charlie said: & "You're the professor." & Choose the most appropriate meaning of the above utterance from the following options.\newline
(A) Emily was hired as a professor.\newline
\textbf{(B)} Emily knows a lot, but she talks too much. \newline
(C) Emily is not good at graduate studies.\newline
(D) Emily lives in a dormitory.\newline
(E) None of the above. \\
\midrule

German & Als Anna, eine Doktorandin, ausführlich über die Theorie sprach, die sie gestern untersucht hatte, sagte Felix:  & "Du bist ja Professorin." & Wählen Sie die passendste Bedeutung der obigen Äußerung aus den folgenden Aussagen aus.\newline
(A) Anna wurde zur Professorin ernannt.\newline
\textbf{(B)} Anna weiß eine Menge, aber sie redet zu viel. \newline
(C) Anna ist nicht gut im Studium.\newline
(D) Anna wohnt in einem Studentenwohnheim.\newline
(E) Keine der obigen Aussagen ist richtig. \\
\midrule

Korean & \begin{CJK}{UTF8}{nanummj}박사생인 영희가 어제 공부한 이론에 대해 길게 이야기하자 철수가 다음과 같이 말했다.\end{CJK} & \begin{CJK}{UTF8}{nanummj}"네가 교수다."\end{CJK} & \begin{CJK}{UTF8}{nanummj}다음 보기에서 위 발화가 갖는 가장 적절한 의미를 고르세요.\newline
(A) 영희는 교수로 임용되었다. \newline
\textbf{(B)} 영희는 아는 것이 많지만 말이 너무 많다. \newline
(C) 영희는 대학원 공부에 소질이 없다. \newline
(D) 영희는 기숙사에 살고 있다. \newline
(E) 정답 없음.\end{CJK} \\
\midrule

Chinese & \begin{CJK*}{UTF8}{gbsn}当博士生王芳详细讲述她昨天学习的理论时，张伟说：\end{CJK*} & \begin{CJK*}{UTF8}{gbsn}"你是教授吗？"\end{CJK*} & \begin{CJK*}{UTF8}{gbsn}请在以下选项中选择最恰当地表达上述话语含义的选项。\newline
(A) 王芳被任命为教授。\newline
\textbf{(B)} 王芳知道很多，但她说得太多了。 \newline
(C) 王芳不适合读研。\newline
(D) 王芳住在宿舍里。\newline
(E) 没有正确答案。\end{CJK*} \\
\bottomrule

\end{tabularx}
\caption{Multilingual test unit example on the maxim of quality. This example illustrates a violation of Grice's maxim of quality, which requires contributions to be true. Although Charlie refers to Emily as "the professor," he does not literally mean that she holds this academic position, as she is a PhD student. Instead, this utterance uses irony to comment on Emily's detailed and extensive explanation, typical of a professor's depth of knowledge. Therefore, the utterance "You're the professor" acknowledges Emily’s thorough knowledge while subtly critiquing her for possibly providing more information than necessary in casual conversation. Thus, option (B) "Emily knows a lot, but she talks too much." best captures the implied meaning of Charlie's statement.}
\label{tab:qual}
\end{table}


\vspace*{\fill}

\begin{table}[h]
\small
\begin{tabularx}{\textwidth}{@{}cp{0.2\textwidth}p{0.2\textwidth}X@{}}
\toprule
\textbf{Language} & \textbf{Context} & \textbf{Utterance} & \textbf{MCQ} \\
\midrule

English & When Charlie confessed to Emily that he wanted to go out with her, she replied: & "I really like you as a friend, too, but I don't think I'm in the right frame of mind to meet someone right now." & Choose the most appropriate meaning of the above utterance from the following options.\newline
(A) Charlie and Emily have a good personality match.\newline
(B) Emily wants to date Charlie's brother.\newline
\textbf{(C)} Emily doesn't want to go out with Charlie.\newline
(D) There are no friends between men and women. \newline
(E) None of the above. \\
\midrule

German & Als Felix Anna gestand, dass er mit ihr ausgehen wollte, sagte sie ihm:  & "Ich mag dich sehr als Freund, aber ich glaube nicht, dass ich im Moment in der richtigen Stimmung bin, um mit jemandem in einer Beziehung sein." & Wählen Sie die passendste Bedeutung der obigen Äußerung aus den folgenden Aussagen aus.\newline
(A) Felix und Anna passen charakterlich gut zusammen.\newline
(B) Anna will mit Felix' Bruder ausgehen.\newline
\textbf{(C)} Anna will nicht mit Felix ausgehen.\newline
(D) Es gibt keine echte Freundschaft zwischen Männern und Frauen. \newline
(E) Keine der obigen Aussagen ist richtig. \\
\midrule

Korean & \begin{CJK}{UTF8}{nanummj}철수가 영희에게 사귀자고 고백하자 영희가 다음과 같이 말했다.\end{CJK} & \begin{CJK}{UTF8}{nanummj}"나도 너를 친구로서 정말 좋아하지만 내가 지금 사람을 만날 만한 마음의 여유가 없는 것 같아."\end{CJK} & \begin{CJK}{UTF8}{nanummj}다음 보기에서 위 발화가 갖는 가장 적절한 의미를 고르세요. \newline
(A) 철수와 영희는 성격이 잘 맞는다.\newline
(B) 영희는 철수의 친오빠와 사귀고 싶다. \newline
\textbf{(C)} 철수와 사귀고 싶지 않다. \newline
(D) 남자와 여자 사이에 친구란 없다. \newline
(E) 정답 없음.\end{CJK} \\
\midrule

Chinese & \begin{CJK*}{UTF8}{gbsn}当张伟向王芳表白，王芳说：\end{CJK*} & \begin{CJK*}{UTF8}{gbsn}"作为朋友我真的很喜欢你，但是我现在状态不适合和别人在一起。"\end{CJK*} & \begin{CJK*}{UTF8}{gbsn}请在以下选项中选择最恰当地表达上述话语含义的选项。\newline
(A) 张伟和王芳性格很合得来。\newline
(B) 王芳想和张伟的哥哥约会。\newline
\textbf{(C)} 王芳不想和张伟谈恋爱。\newline
(D) 男女之间没有朋友。 \newline
(E) 没有正确答案。\end{CJK*} \\
\bottomrule

\end{tabularx}
\caption{Multilingual test unit example on the maxim of manner. Emily’s response to Charlie’s confession is a classic example of violating Grice's maxim of manner, which advocates for clarity and brevity in communication. Instead of a direct answer, Emily’s reply is ambiguously structured, suggesting a rejection without explicitly stating one. This ambiguity is strategic, preserving social harmony while conveying her feelings indirectly. Given the content and context of the conversation, options (A), (B), and (D) do not align with the information provided. Emily emphasizes her current emotional state and her appreciation of their friendship as reasons for not pursuing a romantic relationship, which implicitly suggests she does not wish to date Charlie. Thus, option (C) "Emily doesn't want to go out with Charlie" captures the underlying implication of her response most accurately.}
\label{tab:man}
\end{table}


\vspace*{\fill}

\begin{table}[h]
\small
\begin{tabularx}{\textwidth}{@{}cp{0.2\textwidth}p{0.1\textwidth}X@{}}
\toprule
\textbf{Language} & \textbf{Context} & \textbf{Utterance} & \textbf{MCQ} \\
\midrule

English & Emily and Charlie are working on a writing assignment from class. Emily asks Charlie when the writing assignment is due, and Charlie replies: & "It's due next Thursday." & Choose the most appropriate meaning of the above utterance from the following options.\newline
(A) Charlie is asking Emily for help.\newline
(B) Charlie is not confident in English and wants to postpone the writing assignment.\newline
(C) Charlie wants to finish the writing assignment today.\newline
\textbf{(D)} The writing assignment is due next Thursday.\newline
(E) None of the above. \\
\midrule

German & Anna und Felix arbeiteten an einer schriftlichen Aufgabe aus ihrem Unterricht. Anna fragte Felix, wann die Schreibaufgabe fällig sei, und Felix antwortete:& "Sie ist nächsten Donnerstag fällig." & Wählen Sie die passendste Bedeutung der obigen Äußerung aus den folgenden Aussagen aus.\newline
(A) Er bittet Anna um Hilfe.\newline
(B) Felix ist unsicher in Englisch und möchte die Schreibaufgabe verschieben.\newline
(C) Er möchte die schriftliche Aufgabe sofort fertigstellen.\newline
\textbf{(D)} Die Schreibaufgabe soll bis zum nächsten Donnerstag fertig sein.\newline
(E) Keine der obigen Aussagen ist richtig. \\
\midrule

Korean & \begin{CJK}{UTF8}{nanummj}영희와 철수는 수업에서 나온 글쓰기 과제를 하고 있다. 영희가 철수에게 글쓰기 과제 마감일이 언제인지 묻자, 철수가 다음과 같이 대답했다.\end{CJK} & \begin{CJK}{UTF8}{nanummj}"다음주 목요일까지 제출해야 해."\end{CJK} & \begin{CJK}{UTF8}{nanummj}다음 보기에서 위 발화가 갖는 가장 적절한 의미를 고르세요.\newline
(A) 철수는 영희에게 도움을 요청하는 중이다.\newline
(B) 철수는 영어에 자신이 없어서 글쓰기 과제를 미루고 싶다.\newline
(C) 철수는 오늘 글쓰기 과제를 끝내려고 한다.\newline
\textbf{(D)} 글쓰기 과제 마감일이 다음주 목요일이다.\newline
(E) 정답 없음.\end{CJK} \\
\midrule

Chinese & \begin{CJK*}{UTF8}{gbsn}王芳和张伟正在完成课堂上的写作任务。王芳问张伟什么时候交写作作业，张伟回答说：\end{CJK*} & \begin{CJK*}{UTF8}{gbsn}"下周四前得交上去。"\end{CJK*} & \begin{CJK*}{UTF8}{gbsn}请在以下选项中选择最恰当地表达上述话语含义的选项。\newline
(A) 张伟在向王芳寻求帮助。\newline
(B) 张伟对英语没有信心，想推迟写作任务。\newline
(C) 张伟想在今天完成写作任务。\newline
\textbf{(D)} 下周四之前要交写作作业。\newline
(E) 没有正确答案。\end{CJK*} \\
\bottomrule

\end{tabularx}
\caption{Multilingual test unit example on the category of literal interpretation. Charlie's reply is a direct answer to Emily's question about the deadline. His utterance does not trigger any implications based on the violation of Grice's maxims. It straightforwardly indicates that the due date is next Thursday. Therefore, option (D) "The writing assignment is due next Thursday" is the most appropriate meaning.}
\label{tab:lit}
\end{table}


\vspace*{\fill}

\begin{table}[h]
\small
\begin{tabularx}{\textwidth}{@{}cp{0.2\textwidth}p{0.1\textwidth}X@{}}
\toprule
\textbf{Language} & \textbf{Context} & \textbf{Utterance} & \textbf{MCQ} \\
\midrule

English & Emily saw Charlie's brother in a family photo and asked Charlie how old his brother was, to which he replied: & "He's 28." & Choose the most appropriate meaning of the above utterance from the following options.\newline
(A) Charlie does not know his brother's age.\newline
(B) Charlie's brother is not in college.\newline
(C) Charlie doesn't have a brother.\newline
(D) Charlie's brother is unemployed.\newline
\textbf{(E)} None of the above. \\
\midrule

German & Anna sah Felix' Bruder auf einem Familienfoto und fragte ihn, wie alt er sei, woraufhin Felix antwortete:  & "Er ist 28." & Wählen Sie die passendste Bedeutung der obigen Äußerung aus den folgenden Aussagen aus.\newline
(A) Felix weiß nicht, wie alt sein Bruder ist.\newline
(B) Felix' Bruder geht nicht auf eine Universität.\newline
(C) Felix hat keinen Bruder.\newline
(D) Felix' Bruder ist arbeitslos.\newline
\textbf{(E)} Keine der obigen Aussagen ist richtig. \\
\midrule

Korean & \begin{CJK}{UTF8}{nanummj}영희는 철수의 가족사진에서 그의 동생을 보았고, 동생의 나이를 물었다. 이에 철수는 다음과 같이 대답했다.\end{CJK} & \begin{CJK}{UTF8}{nanummj}"28살이야."\end{CJK} & \begin{CJK}{UTF8}{nanummj}다음 보기에서 위 발화가 갖는 가장 적절한 의미를 고르세요.\newline
(A) 철수는 동생 나이를 알지 못한다.\newline
(B) 철수의 동생은 대학생이 아니다.\newline
(C) 철수는 동생이 없다.\newline
(D) 철수의 동생은 무직이다.\newline
\textbf{(E)} 정답 없음.\end{CJK} \\
\midrule

Chinese & \begin{CJK*}{UTF8}{gbsn}王芳在一张全家福照片上看到了张伟的弟弟，并问他几岁了，张伟回答说：\end{CJK*} & \begin{CJK*}{UTF8}{gbsn}"他 28 岁。"\end{CJK*} & \begin{CJK*}{UTF8}{gbsn}请在以下选项中选择最恰当地表达上述话语含义的选项。\newline
(A) 张伟不知道他的弟弟是几岁。\newline
(B) 张伟的弟弟不是大学生。\newline
(C) 张伟没有弟弟。\newline
(D) 张伟的弟弟失业了。\newline
\textbf{(E)} 没有正确答案。\end{CJK*} \\
\bottomrule

\end{tabularx}
\caption{Multilingual test unit example without correct answer. Charlie's reply to Emily's question about his brother's age is straightforward and direct, with no implications based on the violation of Grice's maxims. His response should thus be interpreted as literal meaning: Charlie's brother is 28 years old. Since none of the options (A) to (D) accurately reflect this literal expression, each introducing an unrelated assumption, the correct answer is (E) "None of the above."}
\label{tab:noans}
\end{table}

\clearpage

\section{Prompt Demonstration and Inter-Rater Agreement Analysis}
\label{sec:append_b}

\begin{table}[h]
\centering
\begin{tabular}{p{15.5cm}}
\toprule
\textbf{Prompt} \\ \midrule
\begin{tabular}[c]{@{}l@{}}While visiting Charlie's house, Emily saw a large pile of oranges in the kitchen and asked why there \\were so many. Charlie responded: \textit{(context)}\\ "My uncle lives in Florida." \textit{(statement)}\\ \\ Choose the most appropriate meaning of the above utterance from the following options. \textit{(MCQ)}\\ (A) Charlie's uncle sent the oranges.\\ (B) Charlie's uncle resides in Florida.\\ (C) People in Florida do not like oranges.\\ (D) Charlie's uncle lives in a rural house.\\ (E) None of the above.\end{tabular} \\ \bottomrule
\end{tabular}
\caption{Example of the prompt using a test unit from our suite. It illustrates how the actual prompt is structured into a context and a corresponding statement followed by an MCQ with options. The words with parentheses are for clarification and are not part of the actual prompt.}
\label{tab:prompt}
\end{table}

\vspace*{\fill}

\begin{table}[h]
\centering
\begin{tabular}{@{}llcccc@{}}
\toprule
 &  & \textbf{English} & \textbf{German} & \textbf{Korean} & \textbf{Chinese} \\ \midrule
 & GPT-4 & 0.87 & 0.86 & \underline{0.70} & 0.90 \\
 & GPT-3.5 & 0.86 & 0.85 & 0.86 & 0.88 \\
 & Claude3-Opus & 0.92 & 0.96 & 0.94 & 0.86 \\
 & Claude3-Sonnet & 0.93 & 0.96 & 0.85 & 0.90 \\
Proprietary & Claude3-Haiku & 0.95 & 0.95 & 0.90 & 0.91 \\
 & Mistral-Large & 0.91 & 0.95 & 0.88 & 0.89 \\
 & Mistral-Medium & 0.90 & 0.90 & 0.94 & 0.94 \\
 & Mistral-Small & 0.80 & 0.84 & 0.84 & 0.85 \\ \midrule
 & Llama3-8B & 0.86 & 0.91 & 0.90 & 0.90 \\
 & Llama2-13B & 0.86 & 0.89 & \underline{0.56} & 0.81 \\
 & Llama2-7B & 0.88 & 0.86 & 0.87 & 0.92 \\
Open-Source & Gemma-7B & 0.97 & 0.99 & 0.96 & 0.97 \\
 & Solar-10.7B & 0.94 & 0.92 & 0.94 & 0.94 \\
 & Qwen-14B & 0.96 & 0.95 & \underline{0.69} & 0.95 \\
 & Qwen-7B & 0.96 & 0.97 & 0.95 & 0.91 \\ \bottomrule
\end{tabular}
\caption{Fleiss' Kappa values representing inter-rater agreement across three trials on the MultiPragEval test suite for four languages. Most models demonstrate high Kappa values (above 0.80), indicating strong agreement across trials. However, models such as GPT-4, Llama2-13B, and Qwen-14B exhibit moderate agreement in generating Korean responses (0.56 to 0.70), suggesting some variability in their performance across the different trials.}

\end{table}

\vspace*{\fill}

\clearpage

\section{Score Tables}
\label{sec:scores}

\vspace*{\fill}

\begin{table}[h]
\centering
\begin{tabular}{@{}llrrrrr@{}}
\toprule
 &  & \multicolumn{5}{c}{\textbf{German}} \\ \midrule
 &  & \multicolumn{1}{l}{Quan.} & \multicolumn{1}{l}{Qual.} & \multicolumn{1}{l}{Rel.} & \multicolumn{1}{l}{Man.} & \multicolumn{1}{l}{\textbf{Avg.}} \\ \midrule
 & GPT-4 & \cellcolor[HTML]{DAE8FC}70.56 & \cellcolor[HTML]{34CDF9}76.67 & \cellcolor[HTML]{3399FF}77.22 & \cellcolor[HTML]{ECF4FF}65.56 & \textbf{72.50} \\
 & GPT-3.5 & \cellcolor[HTML]{3399FF}58.89 & \cellcolor[HTML]{DAE8FC}51.67 & \cellcolor[HTML]{34CDF9}53.89 & \cellcolor[HTML]{ECF4FF}47.22 & 52.92 \\
 & Claude-Opus & \cellcolor[HTML]{34CDF9}85.56 & \cellcolor[HTML]{3399FF}87.78 & \cellcolor[HTML]{DAE8FC}85.00 & \cellcolor[HTML]{ECF4FF}72.78 & \textbf{82.78} \\
 & Claude-Sonnet & \cellcolor[HTML]{DAE8FC}53.89 & \cellcolor[HTML]{3399FF}70.00 & \cellcolor[HTML]{34CDF9}66.11 & \cellcolor[HTML]{ECF4FF}50.56 & 60.14 \\
 & Claude-Haiku & \cellcolor[HTML]{ECF4FF}36.67 & \cellcolor[HTML]{34CDF9}51.67 & \cellcolor[HTML]{3399FF}52.78 & \cellcolor[HTML]{DAE8FC}39.44 & 45.14 \\
 & Mistral-Large & \cellcolor[HTML]{DAE8FC}60.00 & \cellcolor[HTML]{34CDF9}70.00 & \cellcolor[HTML]{3399FF}73.33 & \cellcolor[HTML]{ECF4FF}51.67 & 63.75 \\
 & Mistral-Medium & \cellcolor[HTML]{DAE8FC}47.22 & \cellcolor[HTML]{3399FF}68.89 & \cellcolor[HTML]{34CDF9}56.11 & \cellcolor[HTML]{ECF4FF}42.22 & 53.61 \\
\multirow{-8}{*}{Proprietary} & Mistral-Small & \cellcolor[HTML]{DAE8FC}50.56 & \cellcolor[HTML]{34CDF9}53.33 & \cellcolor[HTML]{3399FF}58.89 & \cellcolor[HTML]{ECF4FF}41.67 & 51.11 \\ \midrule
 & Llama3-8B & \cellcolor[HTML]{ECF4FF}35.56 & \cellcolor[HTML]{34CDF9}40.00 & \cellcolor[HTML]{3399FF}46.67 & \cellcolor[HTML]{DAE8FC}37.78 & 40.00 \\
 & Llama2-13B & \cellcolor[HTML]{3399FF}20.00 & \cellcolor[HTML]{ECF4FF}13.33 & \cellcolor[HTML]{DAE8FC}15.00 & \cellcolor[HTML]{34CDF9}17.22 & 16.39 \\
 & Llama2-7B & \cellcolor[HTML]{3399FF}5.56 & \cellcolor[HTML]{DAE8FC}3.89 & \cellcolor[HTML]{ECF4FF}3.33 & \cellcolor[HTML]{34CDF9}5.00 & 4.44 \\
 & Gemma-7B & \cellcolor[HTML]{34CDF9}29.44 & \cellcolor[HTML]{DAE8FC}23.89 & \cellcolor[HTML]{3399FF}35.00 & \cellcolor[HTML]{ECF4FF}20.56 & 27.22 \\
 & Solar-10B & \cellcolor[HTML]{DAE8FC}56.67 & \cellcolor[HTML]{34CDF9}59.44 & \cellcolor[HTML]{3399FF}62.78 & \cellcolor[HTML]{ECF4FF}43.89 & \textbf{55.69} \\
 & Qwen-14B & \cellcolor[HTML]{3399FF}53.89 & \cellcolor[HTML]{DAE8FC}38.89 & \cellcolor[HTML]{34CDF9}45.56 & \cellcolor[HTML]{ECF4FF}33.89 & 43.06 \\
\multirow{-7}{*}{Open-Source} & Qwen-7B & \cellcolor[HTML]{3399FF}45.56 & \cellcolor[HTML]{DAE8FC}37.78 & \cellcolor[HTML]{34CDF9}41.11 & \cellcolor[HTML]{ECF4FF}33.33 & 39.44 \\ \bottomrule
\end{tabular}
\caption{Performance scores on the MultiPragEval test suite across four maxims with overall averages for German. While the maxim of manner generally shows the lowest scores, high scores are more evenly distributed across the other three maxims.}

\end{table}

\vspace*{\fill}

\begin{table}[h]
\centering
\begin{tabular}{@{}llrrrrr@{}}
\toprule
 &  & \multicolumn{5}{c}{\textbf{Korean}} \\ \midrule
 &  & \multicolumn{1}{l}{Quan.} & \multicolumn{1}{l}{Qual.} & \multicolumn{1}{l}{Rel.} & \multicolumn{1}{l}{Man.} & \multicolumn{1}{l}{\textbf{Avg.}} \\ \midrule
 & GPT-4 & \cellcolor[HTML]{DAE8FC}81.67 & \cellcolor[HTML]{3399FF}86.67 & \cellcolor[HTML]{34CDF9}85.56 & \cellcolor[HTML]{ECF4FF}71.11 & \textbf{81.25} \\
 & GPT-3.5 & \cellcolor[HTML]{34CDF9}42.22 & \cellcolor[HTML]{3399FF}47.22 & \cellcolor[HTML]{DAE8FC}37.22 & \cellcolor[HTML]{ECF4FF}28.89 & 38.89 \\
 & Claude-Opus & \cellcolor[HTML]{DAE8FC}86.67 & \cellcolor[HTML]{34CDF9}87.78 & \cellcolor[HTML]{3399FF}93.33 & \cellcolor[HTML]{ECF4FF}80.56 & \textbf{87.08} \\
 & Claude-Sonnet & \cellcolor[HTML]{DAE8FC}58.89 & \cellcolor[HTML]{3399FF}74.44 & \cellcolor[HTML]{34CDF9}67.78 & \cellcolor[HTML]{ECF4FF}52.22 & 63.33 \\
 & Claude-Haiku & \cellcolor[HTML]{DAE8FC}37.22 & \cellcolor[HTML]{3399FF}49.44 & \cellcolor[HTML]{34CDF9}37.78 & \cellcolor[HTML]{ECF4FF}29.44 & 38.47 \\
 & Mistral-Large & \cellcolor[HTML]{DAE8FC}67.78 & \cellcolor[HTML]{34CDF9}68.33 & \cellcolor[HTML]{3399FF}74.44 & \cellcolor[HTML]{ECF4FF}51.67 & 65.56 \\
 & Mistral-Medium & \cellcolor[HTML]{3399FF}59.44 & \cellcolor[HTML]{DAE8FC}51.11 & \cellcolor[HTML]{34CDF9}53.89 & \cellcolor[HTML]{ECF4FF}47.22 & 52.92 \\
\multirow{-8}{*}{Proprietary} & Mistral-Small & \cellcolor[HTML]{DAE8FC}41.11 & \cellcolor[HTML]{3399FF}52.22 & \cellcolor[HTML]{34CDF9}42.78 & \cellcolor[HTML]{ECF4FF}25.56 & 40.42 \\ \midrule
 & Llama3-8B & \cellcolor[HTML]{34CDF9}34.44 & \cellcolor[HTML]{3399FF}39.44 & \cellcolor[HTML]{DAE8FC}31.11 & \cellcolor[HTML]{ECF4FF}25.00 & 32.50 \\
 & Llama2-13B & \cellcolor[HTML]{34CDF9}45.00 & \cellcolor[HTML]{3399FF}61.11 & \cellcolor[HTML]{DAE8FC}42.22 & \cellcolor[HTML]{ECF4FF}41.67 & 47.50 \\
 & Llama2-7B & \cellcolor[HTML]{3399FF}5.56 & \cellcolor[HTML]{34CDF9}5.00 & \cellcolor[HTML]{ECF4FF}0.00 & \cellcolor[HTML]{DAE8FC}1.67 & 3.06 \\
 & Gemma-7B & \cellcolor[HTML]{34CDF9}30.56 & \cellcolor[HTML]{DAE8FC}15.00 & \cellcolor[HTML]{3399FF}25.00 & \cellcolor[HTML]{ECF4FF}12.78 & 20.83 \\
 & Solar-10B & \cellcolor[HTML]{34CDF9}52.78 & \cellcolor[HTML]{DAE8FC}52.22 & \cellcolor[HTML]{3399FF}57.22 & \cellcolor[HTML]{ECF4FF}33.89 & \textbf{49.03} \\
 & Qwen-14B & \cellcolor[HTML]{34CDF9}53.33 & \cellcolor[HTML]{3399FF}58.89 & \cellcolor[HTML]{DAE8FC}44.44 & \cellcolor[HTML]{ECF4FF}42.22 & \textbf{49.72} \\
\multirow{-7}{*}{Open-Source} & Qwen-7B & \cellcolor[HTML]{34CDF9}36.67 & \cellcolor[HTML]{DAE8FC}35.56 & \cellcolor[HTML]{3399FF}38.33 & \cellcolor[HTML]{ECF4FF}30.00 & 35.14 \\ \bottomrule
\end{tabular}
\caption{Performance scores on the MultiPragEval test suite across four maxims with overall averages for Korean. The maxim of quality typically achieves the highest rankings, while the maxim of manner consistently records the lowest scores, reflecting a similar pattern observed in English.}

\end{table}

\clearpage

\vspace*{\fill} 

\begin{table}[h]
\centering
\begin{tabular}{@{}llrrrrr@{}}
\toprule
 &  & \multicolumn{5}{c}{\textbf{Chinese}} \\ \midrule
 &  & \multicolumn{1}{l}{Quan.} & \multicolumn{1}{l}{Qual.} & \multicolumn{1}{l}{Rel.} & \multicolumn{1}{l}{Man.} & \multicolumn{1}{l}{\textbf{Avg.}} \\ \midrule
 & GPT-4 & \cellcolor[HTML]{DAE8FC}59.44 & \cellcolor[HTML]{3399FF}85.00 & \cellcolor[HTML]{34CDF9}72.78 & \cellcolor[HTML]{ECF4FF}57.78 & \textbf{68.75} \\
 & GPT-3.5 & \cellcolor[HTML]{3399FF}47.22 & \cellcolor[HTML]{DAE8FC}42.22 & \cellcolor[HTML]{34CDF9}43.89 & \cellcolor[HTML]{ECF4FF}41.11 & 43.61 \\
 & Claude-Opus & \cellcolor[HTML]{34CDF9}80.56 & \cellcolor[HTML]{3399FF}82.22 & \cellcolor[HTML]{34CDF9}80.56 & \cellcolor[HTML]{ECF4FF}63.33 & \textbf{76.67} \\
 & Claude-Sonnet & \cellcolor[HTML]{DAE8FC}46.11 & \cellcolor[HTML]{3399FF}63.89 & \cellcolor[HTML]{34CDF9}48.33 & \cellcolor[HTML]{ECF4FF}36.11 & 48.61 \\
 & Claude-Haiku & \cellcolor[HTML]{DAE8FC}40.00 & \cellcolor[HTML]{3399FF}52.78 & \cellcolor[HTML]{34CDF9}40.56 & \cellcolor[HTML]{ECF4FF}30.00 & 40.83 \\
 & Mistral-Large & \cellcolor[HTML]{DAE8FC}47.22 & \cellcolor[HTML]{34CDF9}60.56 & \cellcolor[HTML]{3399FF}66.67 & \cellcolor[HTML]{ECF4FF}44.44 & 54.72 \\
 & Mistral-Medium & \cellcolor[HTML]{34CDF9}43.89 & \cellcolor[HTML]{3399FF}46.67 & \cellcolor[HTML]{DAE8FC}36.67 & \cellcolor[HTML]{ECF4FF}28.33 & 38.89 \\
\multirow{-8}{*}{Proprietary} & Mistral-Small & \cellcolor[HTML]{DAE8FC}35.56 & \cellcolor[HTML]{3399FF}41.11 & \cellcolor[HTML]{34CDF9}39.44 & \cellcolor[HTML]{ECF4FF}18.33 & 33.61 \\ \midrule
 & Llama3-8B & \cellcolor[HTML]{DAE8FC}45.56 & \cellcolor[HTML]{34CDF9}49.44 & \cellcolor[HTML]{3399FF}53.33 & \cellcolor[HTML]{ECF4FF}38.89 & 46.81 \\
 & Llama2-13B & \cellcolor[HTML]{DAE8FC}6.67 & \cellcolor[HTML]{3399FF}12.78 & \cellcolor[HTML]{DAE8FC}3.33 & \cellcolor[HTML]{34CDF9}12.22 & 8.75 \\
 & Llama2-7B & \cellcolor[HTML]{3399FF}7.78 & \cellcolor[HTML]{DAE8FC}3.33 & \cellcolor[HTML]{ECF4FF}0.56 & \cellcolor[HTML]{34CDF9}5.00 & 4.17 \\
 & Gemma-7B & \cellcolor[HTML]{3399FF}29.44 & \cellcolor[HTML]{34CDF9}26.67 & \cellcolor[HTML]{ECF4FF}18.89 & \cellcolor[HTML]{DAE8FC}25.28 & 25.28 \\
 & Solar-10B & \cellcolor[HTML]{34CDF9}49.44 & \cellcolor[HTML]{3399FF}57.78 & \cellcolor[HTML]{DAE8FC}46.67 & \cellcolor[HTML]{ECF4FF}31.67 & 46.39 \\
 & Qwen-14B & \cellcolor[HTML]{34CDF9}51.67 & \cellcolor[HTML]{DAE8FC}47.22 & \cellcolor[HTML]{3399FF}58.89 & \cellcolor[HTML]{ECF4FF}42.22 & \textbf{50.00} \\
\multirow{-7}{*}{Open-Source} & Qwen-7B & \cellcolor[HTML]{DAE8FC}45.00 & \cellcolor[HTML]{3399FF}46.11 & \cellcolor[HTML]{DAE8FC}35.56 & \cellcolor[HTML]{ECF4FF}37.78 & 41.11 \\ \bottomrule
\end{tabular}
\caption{Performance scores on the MultiPragEval test suite across four maxims with overall averages for Chinese. The pattern of rankings mirrors those observed in English and Korean.}

\end{table}

\vspace*{\fill}

\clearpage

\section{Example Responses}

\vspace*{\fill}

\begin{table}[h]
\small
\centering
\begin{tabularx}{\textwidth}{l|X|X|X|X|X}
\hline
\textbf{Id} & \multicolumn{1}{p{1.5cm}|}{7} & \multicolumn{1}{l|}{\textbf{Category}} & \multicolumn{1}{p{3.5cm}|}{Maxim of Quantity} & \multicolumn{1}{l|}{\textbf{Correct Answer}} & \multicolumn{1}{p{1cm}}{A} \\ \hline
\textbf{Question} & \multicolumn{5}{l}{\begin{tabular}[c]{@{}l@{}}\\Charlie and Emily, who are in the same club, got into an argument at a club meeting last week. It's almost \\time for today's meeting and Charlie hasn't shown up. Michael, a friend from the club, asks Emily why he \\hasn't shown up. Emily tells him as follows:\\ \\ "Either he comes or he doesn't."\\ \\ Choose the most appropriate meaning of the above utterance from the following options.\\ (A) Emily doesn't care if Charlie comes or not.\\ (B) Charlie may or may not come. \\ (C) Emily decided to apologize to Charlie.\\ (D) Emily is not a good speaker.\\ (E) None of the above. \\ \\ \end{tabular}} \\ \hline
\textbf{\begin{tabular}[c]{@{}l@{}}Answer by\\ GPT-4\end{tabular}} & \multicolumn{5}{l}{\begin{tabular}[c]{@{}l@{}}\\The most appropriate meaning of Emily's statement "Either he comes or he doesn't" is: \\ (B) Charlie may or may not come.\\ \\ This interpretation is based on the literal meaning of the statement, which indicates uncertainty or \\neutrality regarding Charlie's attendance. It simply states the possibilities without any additional \\emotional or situational context. \\\\ \end{tabular}} \\ \hline
\textbf{\begin{tabular}[c]{@{}l@{}}Answer by \\ Claude3-Opus\end{tabular}} & \multicolumn{5}{l}{\begin{tabular}[c]{@{}l@{}}\\The most appropriate meaning of Emily's utterance "Either he comes or he doesn't" is: \\ (A) Emily doesn't care if Charlie comes or not.\\ \\ Explanation:\\ Emily's statement suggests indifference towards Charlie's attendance at the club meeting. By saying \\"Either he comes or he doesn't," she implies that she is not concerned about whether Charlie shows up \\or not, likely due to their recent argument.\\ \\ The other options are less fitting because:\\ (B) While the statement does suggest uncertainty about Charlie's attendance, it doesn't capture Emily's \\indifference.\\ (C) There is no indication in the given context that Emily has decided to apologize to Charlie.\\ (D) The statement does not reflect on Emily's speaking abilities.\\ (E) Option (A) is the most appropriate interpretation of the given utterance. 
\\\\ \end{tabular}} \\ \hline
\end{tabularx}
\caption{Responses by GPT-4 and Claude3-Opus on a test unit evaluating the Maxim of Quantity in English. While Claude3-Opus captures the implied nuance of indifference in the utterance, GPT-4 focuses on a literal interpretation, leading to an incorrect answer.}

\end{table}

\vspace*{\fill} 
\clearpage

\vspace*{\fill} 

\begin{table}[h]
\small
\centering
\begin{tabularx}{\textwidth}{l|X|X|X|X|X}
\hline
\textbf{Id} & \multicolumn{1}{p{1.5cm}|}{127} & \multicolumn{1}{l|}{\textbf{Category}} & \multicolumn{1}{p{3.5cm}|}{Maxim of Relation} & \multicolumn{1}{l|}{\textbf{Correct Answer}} & \multicolumn{1}{p{1cm}}{A} \\ \hline
\textbf{Question} & \multicolumn{5}{l}{\begin{tabular}[c]{@{}l@{}}\\At a company dinner, a coworker invited Charlie to have a drink, and Charlie said:\\ \\ "I drove here."\\ \\ Choose the most appropriate meaning of the above utterance from the following options.\\ (A) Charlie will not drink alcohol.\\ (B) The car is parked in the parking lot.\\ (C) Charlie plans to drink.\\ (D) Charlie is planning to move.\\ (E) None of the above.\\\\ \end{tabular}} \\ \hline
\textbf{\begin{tabular}[c]{@{}l@{}}Answer by\\ Llama2-13B\end{tabular}} & \multicolumn{5}{l}{\begin{tabular}[c]{@{}l@{}}\\Answer: \\ (C) Charlie plans to drink.\\ \\ Explanation:\\ \\ The statement "I drove here" is often used as a way of saying that one plans to drink alcohol. The speaker \\is indicating that they have arrived at the destination (the company dinner) by driving, and that they are \\prepared to consume alcohol. Therefore, option (C) is the most appropriate meaning of the utterance.\\\\ \end{tabular}} \\ \hline
\textbf{\begin{tabular}[c]{@{}l@{}}Answer by \\ Llama2-7B\end{tabular}} & \multicolumn{5}{l}{\begin{tabular}[c]{@{}l@{}}\\Answer: (C) Charlie plans to drink.\\ \\ Explanation:\\ The sentence "I drove here" is a common way to indicate that the speaker plans to drink alcohol. It implies \\that the speaker arrived at the location by driving, and therefore, \\they are not planning to drink. Therefore, option (C) is the most appropriate meaning of the utterance.\\\\ \end{tabular}} \\ \hline
\end{tabularx}
\caption{Responses by Llama2-13B and Llama2-7B on a test unit evaluating the Maxim of Relation in English. Both models fail to capture the implied meaning of the utterance. While Llama2-7B demonstrates some degree of adequate inference in its reasoning, it still selects the incorrect option.}
\end{table}

\vspace*{\fill} 
\clearpage

\vspace*{\fill} 

\begin{table}[h]
\small
\centering
\begin{tabularx}{\textwidth}{l|X|X|X|X|X}
\hline
\textbf{Id} & \multicolumn{1}{p{1.5cm}|}{75} & \multicolumn{1}{l|}{\textbf{Category}} & \multicolumn{1}{p{3.5cm}|}{Maxim of Quality} & \multicolumn{1}{l|}{\textbf{Correct Answer}} & \multicolumn{1}{p{1cm}}{A} \\ \hline
\textbf{Question} & \multicolumn{5}{l}{\begin{tabular}[c]{@{}l@{}}\\Emily, who saw Charlie come to work on Sunday, said:\\ \\ "He's not even breathing."\\ \\ Choose the most appropriate meaning of the above utterance from the following options.\\ (A) Charlie works hard and diligently.\\ (B) Charlie doesn't breathe when he works.\\ (C) Time passes slowly, as if Charlie is out of breath.\\ (D) Charlie is a bad sleeper.\\ (E) None of the above.\\\\ \end{tabular}} \\ \hline
\textbf{\begin{tabular}[c]{@{}l@{}}Korean Answer \\ by Llama2-13B\end{tabular}} & \multicolumn{5}{l}{\begin{tabular}[c]{@{}l@{}} \\ \begin{CJK}{UTF8}{nanummj}답: ① 철수는 바쁘게 부지런히 일한다.\end{CJK}\\\\ \begin{CJK}{UTF8}{nanummj}설명: 위 발화 "쟤는 숨도 안 쉬어"는 "철수는 바쁘게 부지런히 일한다"라는 의미를 갖고 있습니다. \end{CJK}\\ \begin{CJK}{UTF8}{nanummj}이 발화는 철수가 일요일에 나와서 일하는 것을 본 영희가 철수의 열의와 힘든 일을 직감적으로 표현\end{CJK}\\ \begin{CJK}{UTF8}{nanummj}한 것입니다. \end{CJK} \\\\ \end{tabular}} \\ \hline
\textbf{\begin{tabular}[c]{@{}l@{}}Translation of \\ Korean Answer \end{tabular}} & \multicolumn{5}{l}{\begin{tabular}[c]{@{}l@{}} \\Answer: \begin{CJK}{UTF8}{nanummj}①\end{CJK} Charlie works hard and diligently. \\\\Explanation: The utterance "He's not even breathing." means "Charlie works hard and diligently." This \\utterance is Emily’s instinctive expression of Charlie's enthusiasm and difficulty at work after seeing \\him come to work on Sunday. \\\\ \end{tabular}} \\ \hline
\textbf{\begin{tabular}[c]{@{}l@{}}Chinese Answer \\ by Llama2-13B\end{tabular}} & \multicolumn{5}{l}{\begin{tabular}[c]{@{}l@{}} \\ \begin{CJK*}{UTF8}{gbsn}答：B \end{CJK*} \\\\ \end{tabular}} \\ \hline
\textbf{\begin{tabular}[c]{@{}l@{}}Translation of \\ Chinese Answer \end{tabular}} & \multicolumn{5}{l}{\begin{tabular}[c]{@{}l@{}} \\Answer: B \\\\ \end{tabular}} \\ \hline
\end{tabularx}
\caption{Responses by Llama2-13B on a test unit evaluating the Maxim of Quality in Korean and Chinese. The model demonstrates a significant performance gap between the two languages. While it selects the correct answer and provides a well-reasoned explanation in Korean, it chooses the literal interpretation without any explanation in Chinese.}
\end{table}

\vspace*{\fill} 
\clearpage

\vspace*{\fill} 

\begin{table}[h]
\small
\centering
\begin{tabularx}{\textwidth}{l|X|X|X|X|X}
\hline
\textbf{Id} & \multicolumn{1}{p{1.5cm}|}{169} & \multicolumn{1}{l|}{\textbf{Category}} & \multicolumn{1}{p{3.5cm}|}{Maxim of Relation} & \multicolumn{1}{l|}{\textbf{Correct Answer}} & \multicolumn{1}{p{1cm}}{E} \\ \hline
\textbf{Question} & \multicolumn{5}{l}{\begin{tabular}[c]{@{}l@{}}\\Charlie and Emily walked into a cafe. As Emily was choosing a beverage from the menu, Charlie said:\\ \\ "I have a gift card."\\ \\ Choose the most appropriate meaning of the above utterance from the following options.\\ (A) Charlie wants Emily to give him a gift card.\\ (B) Charlie gave Emily a gift card.\\ (C) Charlie will save the gift card instead of using it.\\ (D) Charlie used to work part-time at a cafe.\\ (E) None of the above. \\\\  \end{tabular}} \\ \hline
\textbf{\begin{tabular}[c]{@{}l@{}}Answer by \\ GPT-4\end{tabular}} & \multicolumn{5}{l}{\begin{tabular}[c]{@{}l@{}}\\The most appropriate meaning of Charlie's utterance "I have a gift card" in the context provided is:\\ \\ (E) None of the above.\\ \\ The correct interpretation, which isn't listed among the options, is likely that Charlie is indicating he has \\a gift card to use at the cafe, possibly suggesting he can pay for the beverages using the gift card. This \\interpretation is not explicitly stated in the options provided.\\\\ \end{tabular}} \\ \hline
\textbf{\begin{tabular}[c]{@{}l@{}}Answer by \\ Claude3-Opus\end{tabular}} & \multicolumn{5}{l}{\begin{tabular}[c]{@{}l@{}}\\The most appropriate meaning of Charlie's utterance "I have a gift card" is:\\ \\ (E) None of the above.\\ \\ In the given context, Charlie mentioning that he has a gift card most likely implies that he intends to use \\it to pay for their purchases at the cafe. This suggests that:\\ \\ - Charlie is offering to treat Emily using his gift card, or\\ - Charlie is informing Emily that they can use the gift card to save money on their purchases.\\ \\ The other options (A, B, C, and D) do not accurately represent the meaning of Charlie's statement in the \\given context.\\\\ \end{tabular}} \\ \hline
\end{tabularx}
\caption{Responses by GPT-4 and Claude3-Opus on a test unit where no correct answer is provided. Both models correctly identify the absence of a correct answer and successfully explain the intended meaning of the utterance in their responses.}
\end{table}

\vspace*{\fill} 
\clearpage

\end{document}